\documentclass[letterpaper, 10 pt, conference]{ieeeconf}  

\IEEEoverridecommandlockouts                              

\usepackage{graphicx} 
\usepackage{multirow}
\usepackage{url}
\usepackage{xcolor}
\usepackage{booktabs}
\usepackage{flushend}
\usepackage{amsmath} 
\usepackage{amssymb}  

\title{\LARGE \bf
FocusGraph: Graph-Structured Frame Selection for Embodied Long Video Question Answering
}

\author{Tatiana Zemskova$^{1,2,*}$, 
Solomon Andryushenko$^{3}$,
Ilya Obrubov$^{4}$, 
Viktoriia Khoruzhaia$^{2}$, \\
Ekaterina Eroshenko$^{2}$,
Ekaterina Derevyanka$^{4}$,
Dmitry Yudin$^{1,2}$
\thanks{$^{1}$ AXXX, $^{2}$ MIRAI, $^{3}$ Yandex, $^{4}$ FusionBrain Lab}
\thanks{$^{*}$Corresponding author: t.s.zemskova@gmail.com}
}

\begin{document}

\maketitle
\thispagestyle{empty}
\pagestyle{empty}

\begin{abstract}
Understanding long videos is crucial for embodied intelligent agents, as their performance depends on effectively accumulating and using long-horizon perceptual memories. Multimodal large language models (MLLMs) are increasingly used for long-video understanding, but their performance degrades and inference time increases as more frames are provided. Therefore, selecting informative keyframes is essential for efficient question answering over long videos. In this work, we develop FocusGraph, a framework for keyframe selection in egocentric long-video question answering. It includes a lightweight Scene-Graph LLM Selector that identifies query-relevant clips from compact graph-based captions, avoiding the need to process raw frame sequences at question time. From these clips, we extract keyframes using Patch-wise Sparse-Flow Retention (PSFR), an offline program-evolved method with no learned parameters at inference time, before passing them to an MLLM for answer generation. FocusGraph achieves state-of-the-art performance on FindingDory and HourVideo while reducing question-time inference cost compared with existing approaches.

\end{abstract}

\section{INTRODUCTION}
Embodied agents operate over extended periods, so their decisions may depend
not only on the current view but also on objects, places, and interactions
observed much earlier~\cite{long2025seeing,anwar2025remembr,fan2025embodied}.
Long-video question answering (LVQA) captures this requirement by asking an
agent to retrieve and combine evidence from its visual history. Such evidence
can be sparse, repetitive, and separated by minutes
~\cite{chandrasegaran2024hourvideo,grauman2022ego4d,yadav2025findingdory}.

The relevant moments are also not necessarily the most visually distinctive.
A question may refer to an object through an earlier interaction or ask which
event preceded another, requiring object recognition, relational reasoning,
and temporal ordering. Camera motion, occlusion, and viewpoint changes further
make repeated observations difficult to associate. Embodied LVQA therefore
requires a structured long-term memory that preserves both visual evidence and
relationships across time.

Directly providing more frames to an MLLM increases cost and can dilute relevant
evidence~\cite{wang2025test,wu2024longvideobench}. Existing methods address this
by compressing visual tokens~\cite{li2024videochat,shenlongvu} or selecting
query-relevant frames
~\cite{zhang2025q,tang2025adaptive,li2025maxinfo,DATE,tang2025tspo}. Compression
reduces context length but may remove details needed for the answer. Frame
selection preserves original observations, but must locate sparse evidence in
a long and redundant sequence.

These alternatives conflate two distinct decisions: determining which events
are semantically relevant and choosing which frames best represent them.
Framewise relevance scores may miss interactions and event order, whereas
global visual diversity can favor distinct but irrelevant scenes. We therefore
hypothesize that the two decisions should be resolved at different scales:
relational memory first locates relevant intervals, after which lightweight
visual cues remove redundancy within them.

\begin{figure}[t]
    \centering
    \includegraphics[width=1.0\linewidth]{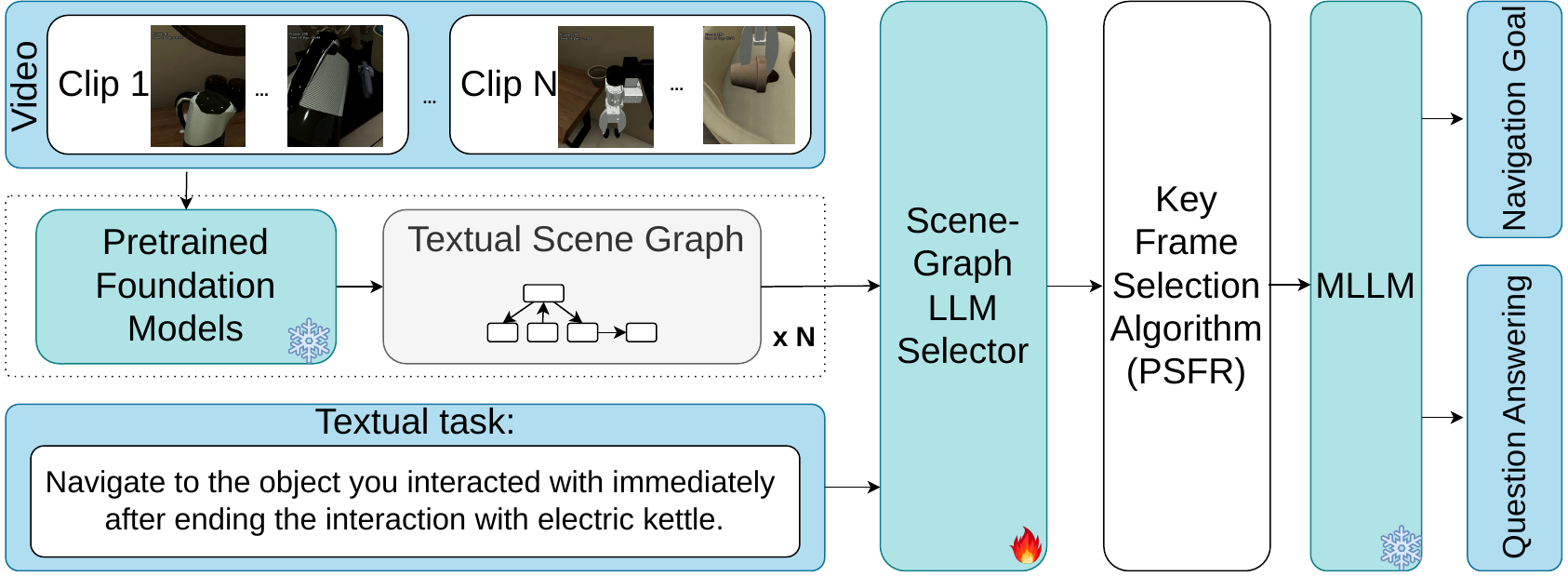}
    \caption{FocusGraph retrieves relevant intervals from a reusable scene-graph memory and uses PSFR to select non-redundant original frames for answer generation.}
    \label{fig:focusgraph-ga}
\end{figure}

FocusGraph implements this separation in two stages. A graph selector reasons
over chronological object-centric triplets to identify relevant clips; Patch-wise Sparse-Flow Retention (PSFR)
then selects non-redundant original frames using sparse flow and appearance
cues. PSFR is evolved offline but is deterministic and parameter-free at
inference. FocusGraph achieves state-of-the-art overall accuracy on
FindingDory~\cite{yadav2025findingdory} and
HourVideo~\cite{chandrasegaran2024hourvideo} with efficient question-dependent
selection after reusable video indexing.

In summary, our contributions are the following:

\begin{itemize}

\item We introduce FocusGraph, a coarse-to-fine framework that separates semantic clip retrieval from visual redundancy reduction for embodied LVQA.
\item We propose a learnable long-video representation that maps temporally ordered object-relation triplets into the LLM embedding space for grounded retrieval.
\item We propose PSFR, an offline-evolved deterministic keyframe selector that combines sparse optical flow with appearance cues and requires no learned parameters at inference.
\item We demonstrate state-of-the-art overall accuracy and efficient question-dependent frame selection on FindingDory and HourVideo.
\end{itemize}

The source code is available at \url{https://github.com/CognitiveAISystems/FocusGraph}.

\section{RELATED WORK}

\textbf{Efficient long-video reasoning.}
Existing methods reduce the visual context through frame panels
~\cite{Panels}, token pruning~\cite{AIM}, codec signals
~\cite{Plug-and-Play}, or textual and temporal annotations
~\cite{RoboAnnotatorX,DATE}. Frame-selection methods are either query-aware,
such as AKS~\cite{tang2025adaptive} and Q-Frame~\cite{zhang2025q}, or
query-agnostic, optimizing motion, diversity, or redundancy
~\cite{lisee,li2025maxinfo,zhu2025focus}. Memory-based approaches retrieve
evidence iteratively~\cite{xie2024embodied,fan2025agentic,
anwar2025remembr,guo2025logic}, whereas learned selectors directly predict
frames or temporal intervals~\cite{yao2025generative,xu2025viarl,
wang2025videoitg}. FocusGraph combines the complementary properties of these
approaches: it performs query-dependent relational retrieval over a reusable
video memory and applies lightweight visual selection only within the retrieved
intervals.

\textbf{Scene graphs for video QA.}
Dynamic and action-centric scene graphs support compositional reasoning over
objects and interactions~\cite{mao2022dynamic,rodin2024action}. Recent methods
extend this idea to compressed graph reasoning~\cite{ling2025accelerating},
temporal triplets~\cite{zong2025structuring}, and textual graph sequences
~\cite{linok2025dygenc}. However, densely representing long videos with
per-frame graphs can produce long and redundant graph sequences under
egocentric camera motion. FocusGraph instead consolidates relations at the clip
level and uses the resulting chronological graph as a question-independent
temporal index. Unlike text-only QA pipelines, the graph is used for retrieval
rather than final answer generation: after localization, the downstream MLLM
receives selected original frames.

\section{METHOD}

\begin{figure*}[t]
    \centering
    \includegraphics[width=1.0\linewidth]{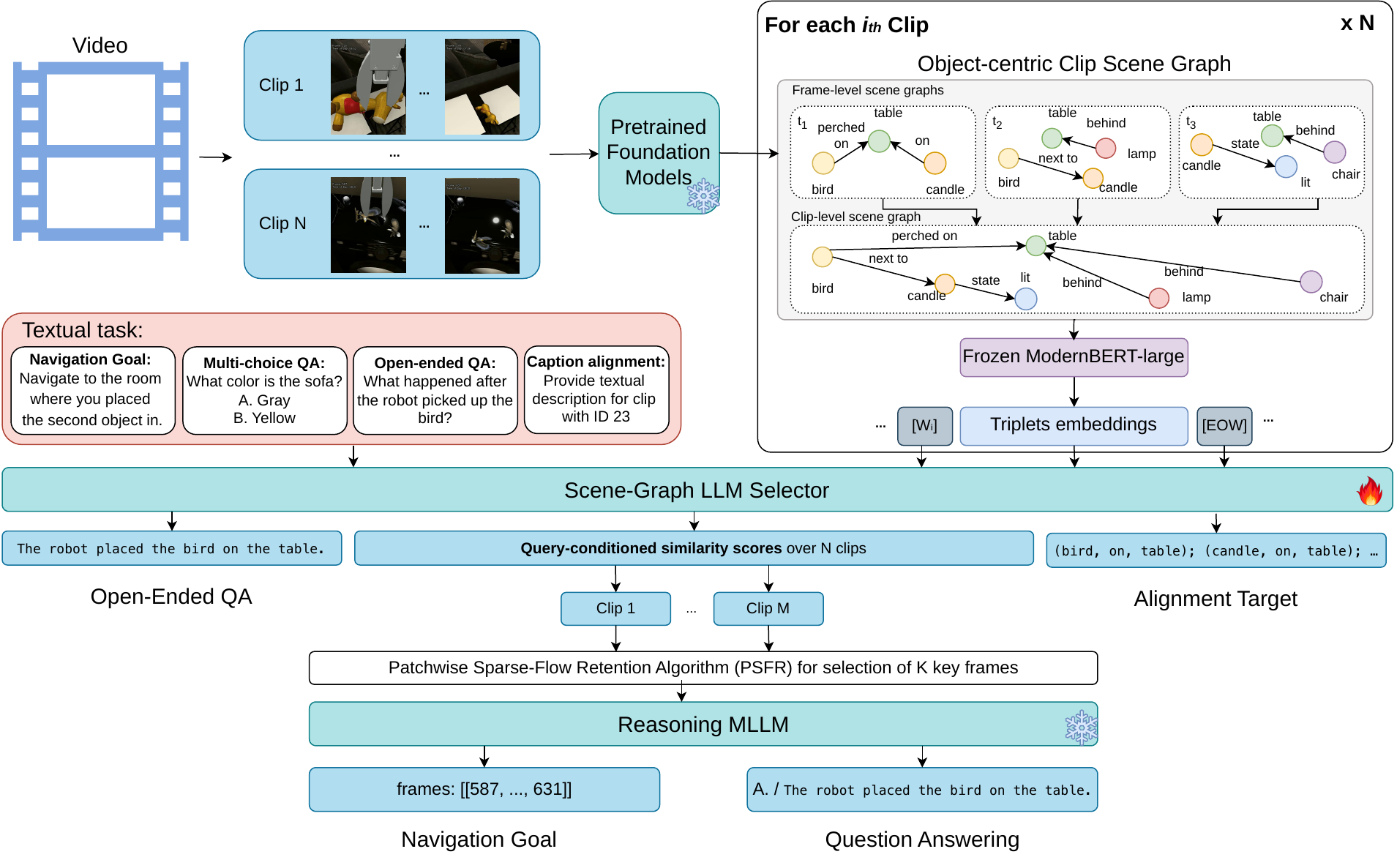}
    \caption{\textbf{FocusGraph overview.}
  FocusGraph partitions an egocentric video into temporal clips represented by object-centric scene graphs generated with the SG-Ego pipeline~\cite{pistilli2026learning}. The graph triplets are
  encoded into a compact chronological memory and processed jointly with the question by Scene-Graph LLM Selector, which generates a temporally grounded answer and identifies relevant video intervals. PSFR then
  selects up to $K$ informative, non-redundant frames for final answer generation by a downstream MLLM.
    }
    \label{fig:focusgraph-method}
\end{figure*}

\textbf{Overview.}
FocusGraph treats a scene-graph sequence as a reusable index into a long video:
the graph provides global temporal context, while the original frames preserve
visual details needed for the final answer. As shown in
Fig.~\ref{fig:focusgraph-method}, we divide the video into clips and construct a
question-independent graph memory with SG-Ego~\cite{pistilli2026learning}. The
Scene-Graph LLM Selector reads the complete memory with the question and, using
the grounding mechanism of GeLM~\cite{chen2025grounded}, identifies relevant
clips. PSFR then selects up to $K$ non-redundant frames from those clips for a
downstream MLLM. The selector is trained to answer and temporally ground
questions jointly, whereas PSFR is optimized separately through program
evolution.

\subsection{Clip-level Scene Graph Construction}
\label{sec:label-construction}

To reduce frame-level ambiguity from motion and occlusion, we divide
$V=\{f_t\}_{t=1}^{T}$ into $N=90$ equal-duration, non-overlapping clips
$\mathcal{C}=\{C_i\}_{i=1}^{N}$ with intervals $\tau_i$. MultiHop-EgoQA
~\cite{chen2025grounded} is sampled at 5 FPS (10 frames per clip), HourVideo
~\cite{chandrasegaran2024hourvideo} at 0.5 FPS, and FindingDory
~\cite{yadav2025findingdory} at its original 1 FPS.

We construct clip graphs with the training-free SG-Ego pipeline
\cite{pistilli2026learning}. SG-Ego first prompts an MLLM (Qwen-3.5-9B~\cite{team2026qwen3} in our experiments) to express spatial and
functional interactions in sampled frames as
$(\text{subject},\text{relation},\text{object})$ triplets. It then grounds the
entities with GroundingDINO~\cite{liu2024grounding} and consolidates frame-level graphs over time by
tracking object instances with SAM2~\cite{ravi2025sam} and DINOv2~\cite{oquab2023dinov2}. Thus, evidence omitted or
occluded in one frame can still be retained at the clip level. We use these
graph-extraction stages without modification.

The resulting clip representation is
\begin{equation}
\mathcal{G}_i=(\mathcal{O}_i,\mathcal{E}_i,\tau_i), \qquad
\mathcal{E}_i=\{(o_s,r,o_t)\},
\end{equation}
where $\mathcal{O}_i$ contains grounded object instances and $\mathcal{E}_i$
contains their directed relations. The chronological sequence
$\{\mathcal{G}_i\}_{i=1}^{N}$ is built once, independently of the question, and
serves as the video memory for the Scene-Graph LLM Selector.

\subsection{Scene-Graph LLM Selector}
\label{sec:llm-selector}

The Scene-Graph LLM Selector replaces the visual tokens of
GeLM~\cite{chen2025grounded} with graph tokens. This is the key compression
step: the LLM can inspect the complete event sequence at low cost, while the
selected clips still point back to the original frames.

For each clip, we serialize and deduplicate its triplets. Frozen
ModernBERT-large~\cite{modernbert} mean-pools each triplet $x_{ij}$, and an MLP
projects it to one LLM-space token,
$\mathbf{z}_{ij}=f_{\mathrm{proj}}\!\left(\phi(x_{ij})\right)$.
All experimental memories fit the 8192-token context after deduplication, so no
temporal truncation or chunking is used.

We preserve clip boundaries and temporal order by surrounding each triplet
sequence with a learned clip-position token and an end-of-clip token,
\begin{equation}
\mathbf{S}_i=[\mathbf{W}_i,\mathbf{z}_{i1},\ldots,
\mathbf{z}_{im_i},\mathbf{EOW}_{\mathrm{i}}].
\end{equation}
Boundary tokens are also retained for empty graphs, so missing graph content
does not collapse the video timeline. The question and the full chronological
sequence $[\mathbf{S}_1,\ldots,\mathbf{S}_N]$ are then processed jointly by the
LLM.

Following GeLM~\cite{chen2025grounded}, tagged answer phrases form queries
$\hat{\mathbf q}_p$ and end-of-clip states form timeline features
$\hat{\mathbf v}_i$. For valid phrases $\mathcal P$, retrieval scores are
\begin{equation}
\begin{aligned}
\ell_{pi}&=\langle\hat{\mathbf q}_p,\hat{\mathbf v}_i\rangle/0.07,
&a_{pi}&=\operatorname{softmax}_i(\ell_{pi}),\\
r_i&=|\mathcal P|^{-1}\!\sum_{p\in\mathcal P}a_{pi},
&s_i&=(r_{i-1}+r_i+r_{i+1})/3,
\end{aligned}
\end{equation}
using zero boundary padding. The grounding head also predicts clip saliency.

\subsection{Training Procedure}
\label{sec:training}

We train the selector in two stages on SG-Ego graphs and temporal annotations
from MultiHop-EgoQA~\cite{chen2025grounded}. First, ModernBERT and the LLM are
frozen while the projector is optimized to make the LLM autoregressively
reconstruct textual triplets; cross-entropy is computed only on reconstruction
tokens.

We then jointly optimize the language backbone, projector, clip tokens, and
grounding head; ModernBERT remains frozen. Target answers contain paired
temporal tags that associate each grounded phrase with one or more evidence
clips. Following GeLM~\cite{chen2025grounded}, training combines three signals:
answer generation, phrase--clip contrastive alignment, and clip saliency,
\begin{equation}
\mathcal{L}=\mathcal{L}_{\mathrm{LM}}
+\mathcal{L}_{\mathrm{NCE}}+10\mathcal{L}_{\mathrm{sal}}.
\end{equation}
The language-model loss teaches the tagged answer, the multi-positive NCE loss
aligns a phrase with all valid evidence clips, and the weighted saliency loss
handles the strong imbalance between relevant and background clips. This joint
objective lets the selector learn what to answer and where the supporting
evidence occurs.

\subsection{Patch-wise Sparse-Flow Retention (PSFR)}
\label{sec:psfr}

We retain $\mathcal I_\alpha=\{i:s_i>\alpha\max_j s_j\}$, with $\alpha=0.1$
 MultiHop-EgoQA and $0.3$ for FindingDory and HourVideo. These scores are
non-negative; without a valid temporal-tag pair, none are produced. PSFR cues
are cached once on the continuous video $V$. Retrieved clips define
$\mathcal A_q=\bigcup_{i\in\mathcal I_\alpha}\{t:f_t\in C_i\}$, and PSFR returns
$\mathcal F_K=\operatorname{PSFR}_K(V;\mathcal A_q)\subseteq\mathcal A_q$.
Thus, up to $K$ frames are selected jointly, without a per-clip quota. Inspired
by FlowGEBD~\cite{gothe2024s}, PSFR is evolved offline but deterministic and
uses neither learned parameters nor annotations at inference.

PSFR partitions each frame into patches and seeds up to $m$ Shi--Tomasi
corners~\cite{shi1994good} per patch. Pyramidal Lucas--Kanade flow
~\cite{lucas1981iterative} tracks them from the latest reseed time $\tau$. For
patch $j$, let $n_j^\tau$ be the seeded count and $c_j^{t+1}$ the surviving
count. A structural-change event occurs when
\begin{equation}
r_j^{t+1}=c_j^{t+1}/n_j^\tau,\quad
\bigl|\{j:n_j^\tau>0,\ r_j^{t+1}<\tau_r\}\bigr|\ge k_{\min}.
\label{eq:psfr_event_main}
\end{equation}
At an event, the denominators are reseeded; otherwise tracks are refreshed
while the previous denominators are retained.

Corners are detected independently inside each patch, ranked by response, and
deduplicated within radius $\rho$ before the global set is capped at $C$.
Tracking accepts only points for which Lucas--Kanade reports a valid solution;
the surviving set is then topped up with newly detected corners. Retaining the
denominators from the last event is important: continuously resetting them
would make gradual track loss invisible, while never refreshing the point set
would confuse ordinary feature dropout with structural change. Patch-wise
counts also prevent a highly textured region from dominating the event score.
Together, these choices make the signal responsive to viewpoint and scene
changes while remaining robust to local motion and temporary occlusion.

The second stage combines normalized corner count, central-window corners,
Canny edge density~\cite{canny2009computational}, grayscale entropy, and the
low-retention count with HSV-histogram change. It assigns content-aware slots,
prefers high-quality diverse frames, and applies non-maximum suppression. The
selection rule is found by the program-evolution procedure below; the resulting
PSFR program and evaluation harness will be released with the code.

Each cue is robustly normalized between its 5th and 95th percentiles. PSFR
constructs a progress signal from cumulative appearance change and frame
quality, then places up to $K$ slots at equal increments of that signal. When a clip
has little appearance change, the slots interpolate toward uniform temporal
spacing so that static but semantically useful intervals remain covered. A
slot may be snapped to a nearby strong change peak. Within each slot window,
selection balances frame quality, distance from already selected timestamps,
and HSV-histogram dissimilarity. After choosing a frame, non-maximum
suppression reduces the scores of its temporal neighbors. This produces an
ordered set rather than independent top-$K$ frames and prevents a short burst
of camera motion from consuming the entire budget.

\subsection{Program-evolution setup for keyframe selector search}
\label{sec:psfr-selection}
We use OpenEvolve~\cite{openevolve} to search over mappings from cached frame
features to keyframe indices using only the FindingDory training split; validation
questions are excluded. For question $q$, the interface is
$\texttt{select}(S_q,H_q,A_q,K)\to S_q^\star$, where $S_q$ contains five
normalized PSFR cues, $H_q$ contains 432-D HSV histograms, and $A_q$ contains
original-video indices from retrieved clips. Valid programs satisfy
$S_q^\star\subseteq A_q$, $|S_q^\star|\le K$ and a 15-second CPU limit; we use
$K=16$ during evolution. Since $K$ is an input, the same rule is used without
re-optimization at the $K=8$ QA budget; final PSFR returns $\min(K,|A_q|)$
frames for nonempty $A_q$.

With valid ground-truth sets $G_{q,m}$, evidence coverage is
\begin{equation}
\mathrm{Incl}(q)=\min_m\mathbb{I}[|S_q^\star\cap G_{q,m}|>0].
\label{eq:oracle_inclusion}
\end{equation}
Runtime $t_q$ is penalized smoothly by
\begin{equation}
\begin{aligned}
\phi(t_q)&=\exp\!\left(\log(\eta)
\mathrm{clip}(t_q/T_{\max},0,1)^\gamma\right),\\
J&=\frac1N\sum_q\mathrm{Incl}(q)\phi(t_q),
\end{aligned}
\label{eq:combined_objective}
\end{equation}
where $T_{\max}=15$ s, $\eta=0.95$, and $\gamma=1$. Search starts from
uniform sampling and evolves sandboxed implementations of the same interface
across multiple islands. Invalid indices, oversized outputs, execution errors,
and timeouts receive zero score. An archive retains high-performing programs,
while migration between islands preserves diversity. Since visual features are
cached, the search compares selection logic rather than recomputing video
representations. The released code will include the full search configuration,
evaluation harness, and evolved programs.

The cached representation exposes only signals available at deployment. The
score array contains corner abundance, central-region support, edge density,
entropy, and low-retention events; HSV histograms provide an independent
appearance descriptor. Candidate programs may transform these arrays but
cannot inspect annotations. Oracle metrics are used only by the evolutionary
evaluation loop; the deployed deterministic selector has no annotation access.

\section{Experiments}

\subsection{Datasets and Evaluation Protocol}
\textbf{FindingDory}~\cite{yadav2025findingdory} contains mobile-manipulation
episodes recorded in the Habitat simulator using HSSD scenes
~\cite{khanna2024habitat}. We evaluate on the full validation split of 5,870
questions from 100 episodes and on the 211-question development subset formed
by episodes 2, 5, 6, and 9. The full split covers spatial, temporal, and
multi-goal questions, while the smaller subset permits comparison with costly
and agent-based methods. QA performance is reported as exact-match accuracy
overall and for each question category.

FindingDory also provides frame-level supervision for evaluating selection.
Importantly, the target is not a single representative evidence frame. For
each question, the required target contains \emph{all} annotated frames in
which the room or object that is the answer to the question is visible. Thus,
a selector is expected to recover the complete set of answer-bearing frames,
including repeated observations of the same room or object at different
timestamps. We compare selected and target sets with Intersection, Inclusion,
and $F_{\sqrt2}$ under a fixed frame budget; Sec.~IV-E details these metrics.

\textbf{HourVideo}~\cite{chandrasegaran2024hourvideo} comprises hour-long
egocentric videos from Ego4D~\cite{grauman2022ego4d}. We use its standard
development split of 50 videos and 1,182 questions spanning navigation,
perception, visual reasoning, and summarization. We report overall and category
accuracy, averaging FocusGraph over three selection runs.

\textbf{MultiHop-EgoQA}~\cite{chen2025grounded} is used to train the
Scene-Graph LLM Selector and to evaluate joint answering and temporal
grounding, including cases with evidence in disjoint intervals. We report
mIoP, mIoG, IoU@0.3, and mIoU together with semantic answer similarity and an
LLM-based answer score. \textbf{GenS-Videos-150K}
~\cite{yao2025generative} is used only for selector-level evaluation with its
weighted-intersection protocol, testing whether PSFR objectives transfer across
datasets.

Across paired frame-selection comparisons, the downstream prompt, decoding
configuration, answer model, and visual-frame budget are held fixed. For
timing, we separately report reusable window indexing, question-dependent
selection, and the common downstream QA call, avoiding conflation of cached
and per-query costs.

\begin{table*}[ht]
\centering
\tiny
\caption{Performance comparison across frame sampling methods and LLM settings
on FindingDory~\cite{yadav2025findingdory}.}
\begin{tabular}{l l l c cc cc cc cc}
\hline
\multirow{2}{*}{\shortstack{Frame Selection \\ Method}} &
\multirow{2}{*}{\shortstack{Scene \\ Representation}} &
\multirow{2}{*}{QA MLLM} &
\multirow{2}{*}{\shortstack{Frame \\ Number}} &
\multicolumn{2}{c}{\shortstack{All \\ tasks}} &
\multicolumn{2}{c}{\shortstack{Single-Goal \\ Spatial Tasks}} &
\multicolumn{2}{c}{\shortstack{Single-Goal \\ Temporal Tasks}} &
\multicolumn{2}{c}{\shortstack{Multi-Goal \\ Tasks}} \\
\cline{5-12}
 &  &  &  &
 full val. & dev. &
 full val. & dev. &
 full val. & dev. &
 full val. & dev. \\
\hline
Uniform Sampling & image & Gemini-3-Flash~\cite{gemini3flash2025} & 8 & - & 26.42 & - & 31.82 & - & 30.60 & - & 2.1 \\
Uniform Sampling & image & Qwen-2.5-VL-7B~\cite{qwen2.5-VL} & 8 & 12.6 & 18.4 & 16.4 & 22.6 & 11.8 & 19.4 & 2.0 & 3.2 \\
Uniform Sampling & image & GPT-5-Mini~\cite{singh2025openai} & 8 & 15.49 & 24.25 & 20.37 & 29.27 & 14.04 & 27.38 & 2.71 & 2.78 \\
Uniform Sampling & image & Qwen3.8-27B~\cite{qwen3.8-27b} & 8 & 18.12 & \underline{28.97} & 23.38 & 34.57 & 17.08 & \textbf{34.08} & \textbf{3.11} & 2.55 \\
\hline
MaxInfo~\cite{li2025maxinfo} & image & Qwen-2.5-VL-7B~\cite{qwen2.5-VL} & 16 & 13.7 & 17.2 & 17.5 & 19.5 & 13.4 & 22.2 & 1.9 & 2.1 \\
ReMEmbR~\cite{anwar2025remembr} & text & Llama3.1-8B~\cite{grattafiori2024llama} & 7\footnotemark[1] & - & 21.16 & - & 27.2 & - & 19.4 & - & \underline{4.1} \\
ReMEmbR~\cite{anwar2025remembr} & text & GPT-5.2~\cite{singh2025openai} & 51\footnotemark[1] & - & 26.56 & - & 31.1 & - & \underline{32.3} & - & 2.8 \\
GenS~\cite{yao2025generative} & image & Qwen-2.5-VL-7B~\cite{qwen2.5-VL} & 8 & 16.6 & 22.3 & 22.6 & 25.5 & 13.6 & 28.1 & 2.9 & 2.8 \\
ViaRL\cite{xu2025viarl} & image & Qwen-2.5-VL-7B~\cite{qwen2.5-VL} & 8 & 14.9 & 21.4 & 20.8 & 28.0 & 12.3 & 30.0 & 1.7 & 1.2 \\
FocusGraph (ours) & scene graph & Qwen-2.5-VL-7B~\cite{qwen2.5-VL} & 8 & 17.25 & 26.88 & 23.47 & 35.05 & 14.50 & 25.00 & \underline{3.00} & 2.78 \\
FocusGraph (ours) & scene graph & GPT-5-Mini~\cite{singh2025openai} & 8 & \underline{19.08} & 27.29 & \underline{24.54} & \underline{36.27} & \underline{18.33} & 23.21 & 2.74 & 3.94 \\
FocusGraph (ours) & scene graph & Qwen3.8-27B~\cite{qwen3.8-27b} & 8 & \textbf{20.45} & \textbf{29.65} & \textbf{26.11} & \textbf{38.23} & \textbf{20.06} & 26.67 & 2.66 & \textbf{5.93} \\
\hline
\end{tabular}

\label{tab:frame_sampling_results_finding_dory}
\end{table*}

\begin{table}[t]
\centering
\tiny
\caption{Results on HourVideo~\cite{chandrasegaran2024hourvideo}. All methods
use Qwen-2.5-VL-7B~\cite{qwen2.5-VL} except ReMEmbR~\cite{anwar2025remembr},
which uses Llama3.1-8B~\cite{grattafiori2024llama}.}
\resizebox{\linewidth}{!}{%
\begin{tabular}{@{}lcccccc@{}}
\hline
\shortstack{Frame Selection \\ Method} &
\shortstack{Frame \\ No.} &
Overall  &
Nav. &
Perc. &
Reason. &
Summ. \\
\hline
Uniform Sampling & 8 & 29.86 & 10.0 & 33.7 & 27.9 & 37.8  \\
Uniform Sampling & 16 & 31.22 & 15.0 & \underline{36.8} & 27.5 & \textbf{43.9}  \\
\hline
MaxInfo~\cite{li2025maxinfo} & 16 & 29.61 & 17.5 & 33.9 & 26.6 & 40.2  \\
ReMEmbR~\cite{anwar2025remembr} & 8\footnotemark[1] & 24.28 & 17.5 & 24.54 & 27.47 & 0.0 \\
VideoTree~\cite{wang2025videotree} & 8 & 31.4 & 22.5 & 32.4 & 30.4 & 39.0  \\

LVNet~\cite{park2026too} & 8 & 24.79 & \underline{27.50} & 22.72 & 23.93 & 40.24  \\
LifeLongMemory~\cite{wang2023lifelongmemory} & 8 & 29.9 & 17.5 & 31.6 & 28.2 & 42.7  \\
FOCUS~\cite{zhu2025focus} & 8 & 32.4 & \textbf{30.0} & 31.9 & 31.4 & \underline{43.5}  \\
GenS~\cite{yao2025generative} & 8 & \underline{32.9} & \underline{27.5} & 34.5 & \underline{31.8} & 37.8  \\
ViaRL~\cite{xu2025viarl} & 8 & 32.23 & 17.5 & 33.4 & 31.0 & \textbf{43.9} \\
FocusGraph (ours) & 8 & \textbf{35.19} \textpm{} 0.08 & 20.83 \textpm{} 1.44 & \textbf{38.82} \textpm{} 0.60 & \textbf{33.04} \textpm{} 0.56 & 43.09 \textpm{} 0.70 \\
\hline

\end{tabular}%
}
\label{tab:frame_sampling_results_hourvideo}
\end{table}

\begin{table}[t]
\centering
\tiny
\caption{Inference latency on HourVideo~\cite{chandrasegaran2024hourvideo}.}
\setlength{\tabcolsep}{2.5pt}
\begin{tabular}{@{}p{0.43\columnwidth}ccc@{}}
\toprule
Method / stage & Scope & Tok./fr. & Time, ms \\
\midrule
\multicolumn{4}{l}{\textit{Reusable window processing}} \\
Frame-level relation extraction
    & Window & -- & $59180\pm14430$ \\
Object grounding and frame-graph construction
    & Window & -- & $10440\pm6200$ \\
Temporal tracking and graph consolidation
    & Window & -- & $16230\pm9890$ \\
MaxInfo~\cite{li2025maxinfo}
    & Window & -- & 3420 \\
PSFR visual feature extraction (ours)
    & Window & -- & $4530\pm2440$ \\
\cmidrule(lr){1-4}
\textbf{FocusGraph reusable total}
    & Window & --
    & $\mathbf{90380\pm29740}$ \\
\midrule
\multicolumn{4}{l}{\textit{Frame selection only (downstream QA excluded)}} \\
ReMEmbR~\cite{anwar2025remembr}
    & Question & -- & 80000 \\
GenS~\cite{yao2025generative}
    & Question & 16 & 103000 \\
ViaRL~\cite{xu2025viarl}
    & Question & 16 & 2600 \\
Scene-Graph LLM Selector (ours)
    & Question & $<1$ & $1544\pm1022$ \\
FocusGraph: final PSFR sampling (ours)
    & Question & -- & $1.33\pm0.01$ \\
\midrule
\multicolumn{4}{l}{\textit{Common downstream question answering}} \\
Qwen2.5-VL-7B~\cite{qwen2.5-VL} (all methods)
    & Question & -- & $2087\pm113$ \\
\bottomrule
\end{tabular}
\label{tab:inference_speed}
\end{table}

\begin{figure*}[t]
    \centering
    \includegraphics[width=1.0\linewidth]{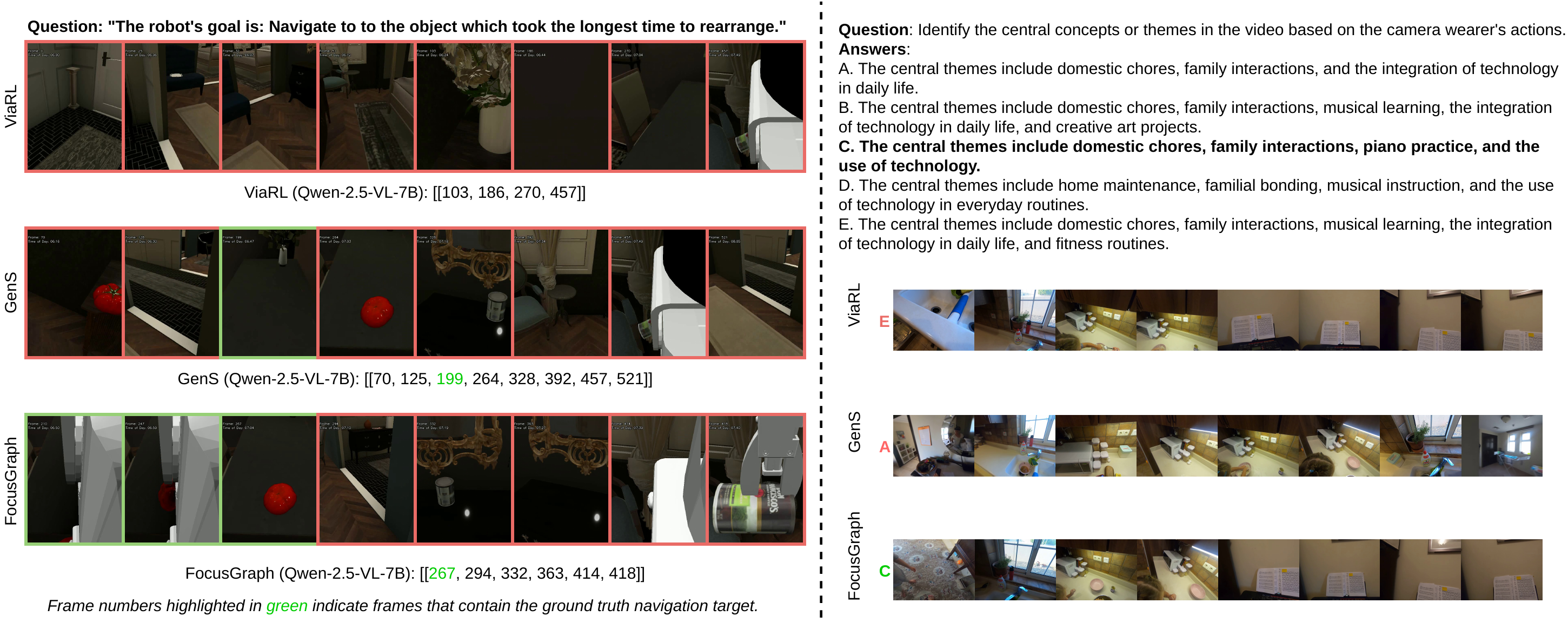}
    \caption{Comparison of frame sampling methods. Left:
    FindingDory~\cite{yadav2025findingdory}. Green-bordered frames are present
    in the ground truth; red-bordered frames are not. Right:
    HourVideo~\cite{chandrasegaran2024hourvideo}. Correct answers are
    highlighted in green, incorrect ones in red.}
    \label{fig:sampling-comparison-findingdory}
\end{figure*}

\begin{table}[t]
\centering
\tiny
\caption{Comparison with published baselines and graph tokenization variants
on the MultiHop-EgoQA~\cite{chen2025grounded} validation split.}
\setlength{\tabcolsep}{2.0pt}
\resizebox{\linewidth}{!}{%
\begin{tabular}{@{}l|cccc|cc@{}}
\toprule
Method &
$\mathrm{mIoP}$ & $\mathrm{mIoG}$ & $\mathrm{IoU@0.3}$ & $\mathrm{mIoU}$ &
\shortstack{Sent.\\Sim.} & Score \\
\midrule
GPT-4o~\cite{hurst2024gpt} (60 frames, z.s.) & 18.9 & 24.4 & 12.0 & 12.2 & 73.7 & \textbf{5.4} \\
InternVL2-8B~\cite{opengvlab2024internvl2} (30 frames, z.s.) & 11.8 & 24.0 & 6.3 & 6.6 & 71.9 & 4.5 \\
LLaVA-NeXT-Video-7B~\cite{zhang2024llavanextvideo} (32 frames, z.s.) & -- & -- & -- & -- & 62.1 & 4.2 \\
TimeChat-7B~\cite{ren2024timechat} (96 frames, z.s.) & 10.2 & 5.6 & 3.0 & 3.6 & 58.9 & 3.3 \\
VTimeLLM-7B~\cite{huang2024vtimellm} (100 frames, z.s.) & 12.4 & 28.2 & 8.8 & 9.2 & 70.5 & 4.3 \\
LLaVA-NeXT-7B~\cite{zhang2024llavanextvideo} $\rightarrow$
Llama-3.1-8B~\cite{grattafiori2024llama} (180 frames, z.s.) & 21.4 & 22.3 & 10.1 & 9.7 & 63.6 & 3.5 \\
GeLM-7B~\cite{chen2025grounded} (180 frames, f.t.) & 23.7 & 41.0 & 18.2 & 16.7 & 75.0 & 4.8 \\
\midrule
\multicolumn{7}{l}{\textit{FocusGraph graph tokenization variants}} \\
Triplets     & \textbf{28.3} & 55.1 & \textbf{27.2} & \textbf{22.4} & \textbf{75.0} & 4.6 \\
Triplets w/o pretraining & 17.6 & \textbf{68.2} & 14.9 & 15.6 & 73.2 & 4.4 \\
Window graph & 23.8 & 50.8 & 21.8 & 18.7 & 73.5 & 4.3 \\
Objects      & 21.1 & 59.3 & 18.8 & 17.7 & 74.6 & 4.5 \\
\bottomrule
\end{tabular}%
}
\label{tab:focusgraph_ablation}
\end{table}

\begin{table}[t]
\centering
\tiny
\caption{Downstream frame-sampling ablation on the
FindingDory~\cite{yadav2025findingdory} full validation split.}
\setlength{\tabcolsep}{2.0pt}
\resizebox{\linewidth}{!}{%
\begin{tabular}{@{}llc|cccc@{}}
\toprule
Clip selection & Frame sampling & \shortstack{Frame\\number} &
\shortstack{All\\tasks} &
\shortstack{Single-goal\\spatial} &
\shortstack{Single-goal\\temporal} &
\shortstack{Multi-goal\\tasks} \\
\midrule
Uniform image baseline & Uniform & 8 & 12.60 & 16.40 & 11.80 & 2.00 \\
Scene-Graph LLM Selector & Uniform & 8 & 15.76 & 21.72 & 12.65 & \textbf{3.15} \\
Scene-Graph LLM Selector & PSFR & 8 & \textbf{17.25} & \textbf{23.47} & \textbf{14.50} & 3.00 \\
\bottomrule
\end{tabular}%
}
\label{tab:focusgraph_sampling_ablation}
\end{table}
\subsection{Baselines}
We compare FocusGraph with query-agnostic, query-conditioned, hierarchical,
and memory-based selectors. Uniform sampling is the query-independent control.
MaxInfo~\cite{li2025maxinfo} promotes visual diversity by maximizing
frame-embedding volume. GenS~\cite{yao2025generative} learns to generate a
compact frame set, whereas ViaRL~\cite{xu2025viarl} learns adaptive temporal
grounding through reinforcement learning. ReMEmbR
~\cite{anwar2025remembr} iteratively retrieves evidence from a textual
spatio-temporal memory.

HourVideo comparisons additionally include the adaptive hierarchy of
VideoTree~\cite{wang2025videotree}, the persistent egocentric memory of
LifeLongMemory~\cite{wang2023lifelongmemory}, the efficient LVQA strategy
LVNet~\cite{park2026too}, and the relevance--redundancy selection of
FOCUS~\cite{zhu2025focus}. On MultiHop-EgoQA, we use the zero-shot MLLMs,
caption-to-LLM pipeline, and fine-tuned GeLM reported with the benchmark
~\cite{chen2025grounded}. Comparisons that share Qwen-2.5-VL-7B
~\cite{qwen2.5-VL} and the same frame budget isolate selection quality; the
additional answer models in Table~\ref{tab:frame_sampling_results_finding_dory}
test compatibility across reasoning backbones.

\subsection{Training Details}
Projector alignment uses 20,000 non-empty SG-Ego~\cite{pistilli2026learning}
clips from MultiHop-EgoQA~\cite{chen2025grounded} training videos. We train the
projector for one pass with AdamW, learning rate
$2\!\times\!10^{-4}$, global batch size 16, and a maximum reconstruction target
of 256 tokens. ModernBERT-large~\cite{modernbert} and
Qwen3.5-9B~\cite{team2026qwen3} are frozen at this stage.

\footnotetext[1]{For ReMEmbR, frame number is the average number of documents retrieved from the database per query.}

For joint answer and grounding training, we use the
MultiHop-EgoQA~\cite{chen2025grounded} training
splits and remove examples whose annotations cannot be represented by the five
supported temporal-tag pairs. We optimize the Qwen backbone, aligned graph
projector, learned clip tokens, and GeLM~\cite{chen2025grounded} grounding head
for 10 epochs with
AdamW. We use a learning rate of $2\!\times\!10^{-5}$ with cosine decay and
$3\%$ warmup, bfloat16 precision, an 8192-token context, and global batch size
64 on 8 NVIDIA A100 80GB GPUs; ModernBERT~\cite{modernbert} remains frozen.
HourVideo~\cite{chandrasegaran2024hourvideo} videos are
excluded from selector training for experiment in Table~\ref{tab:frame_sampling_results_hourvideo}. The downstream QA model is not fine-tuned, so
improvements measure frame selection rather than answer-model adaptation.

\subsection{Comparison with the state-of-the-art methods.}
\label{sec:results}

FocusGraph gives the strongest overall result on both FindingDory splits with
Qwen3.8-27B (Table~\ref{tab:frame_sampling_results_finding_dory}) and improves
uniform sampling across all three tested reasoning backbones. On HourVideo it
achieves the best overall score with only eight frames
(Table~\ref{tab:frame_sampling_results_hourvideo}); gains concentrate on
perception and visual reasoning, whereas broad-coverage navigation and
summarization sometimes favor other samplers. The qualitative examples in
Fig.~\ref{fig:sampling-comparison-findingdory} show that FocusGraph retains
complementary evidence rather than visually similar frames.

The category pattern reflects the two-stage design. Perception and visual
reasoning often depend on a few localizable observations, for which the graph
memory narrows the search before PSFR selects a clear view. Navigation and
summarization require broader coverage, and a fixed eight-frame budget may
discard useful context. FindingDory multi-goal questions expose the same
limitation when evidence lies in several disjoint intervals. FocusGraph is
therefore most effective when relevance is sparse but can be expressed through
objects and their interactions.

Latency results
(see Table~\ref{tab:inference_speed}) show that graph construction is a reusable
one-time cost; question-dependent graph retrieval and final PSFR sampling are
inexpensive and amortize well in multi-query settings.
The latency decomposition separates localization from answer generation:
relation extraction, grounding, and temporal consolidation are shared by every
question about a window, while only compact retrieval, PSFR lookup, and QA
repeat per query. The latency measurements were made on a single RTX 4090.
We use six 180-s HourVideo windows from distinct videos, each represented by
90 frames at 0.5 fps. Stored frames retain their native resolution: three at
$1440\times1080$, two at $1920\times1440$, and one at $2560\times1920$;
relation extraction and object grounding use 384- and 320-pixel inputs,
respectively, whereas PSFR uses the stored resolution. Each reusable stage and
one matched query per video are timed once. The first window--query pair is a
warm-up and excluded, so Table~\ref{tab:inference_speed} reports the mean and
sample standard deviation over five runs; model loading is excluded. PSFR sampling is timed repeatedly: after one warm-up, 500 calls per window are averaged within each window and then across the five measured windows.

\subsection{Ablation studies}
Table~\ref{tab:focusgraph_ablation} compares graph tokenization variants with
published MultiHop-EgoQA baselines. Triplets outperform every baseline on all
four grounding metrics and match the best semantic similarity, although their answer score remains below GPT-4o and GeLM, motivating the use of a reasoning MLLM for downstream QA.
Among our variants, explicit
relations improve precision over object-only and window memories by
disambiguating repeated interactions; removing projector pretraining instead
favors coverage at the expense of precision and answer quality. Thus, stronger
grounding does not automatically translate into a higher judged answer score.

Table~\ref{tab:focusgraph_sampling_ablation} isolates the contributions of the
two frame-selection stages under the same downstream MLLM and frame budget.
The Scene-Graph LLM Selector improves on full-video uniform sampling by
narrowing the search to question-relevant regions. PSFR adds further gains,
especially on spatial and temporal questions. The stages are complementary:
the graph selector determines \emph{where} evidence occurs, while PSFR chooses
\emph{which} non-redundant observations represent it. Its small decrease on
multi-goal questions indicates that aggressive selection can miss distributed
evidence under a fixed budget.

\begin{table}[t]
\centering
\tiny
\caption{
Ablation of keyframe selectors under a fixed budget of $K{=}16$ frames. GenS denotes GenS-Videos-150K.
}
\label{tab:evolved_kfs_objectives}
\begin{tabular}{@{}llcccccc@{}}
\toprule
& &\multicolumn{3}{c}{FindingDory (FD)} & \multicolumn{2}{c}{GenS} & \shortstack{Time (s)} \\
\cmidrule(lr){3-5}\cmidrule(lr){6-7}
Keyframe selector & Opt. &
Inter. & Incl. & $F_{\sqrt{2}}$ &
w-Inter & Incl &
$\mu \pm \sigma$ \\
\midrule

Uniform & - &
0.111 & 0.715 & 0.0136 &
0.470 & \textbf{0.890} &
-- \\

Hand-Crafted & FD &
0.075 & 0.380 & 0.008 &
0.501 & 0.840 &
$0.016 \pm 0.002$ \\

Grid-Search & FD &
0.132 & 0.714 & 0.016 &
0.488 & 0.832 &
$0.017 \pm 0.002$ \\

\hline

Evolved on Inter. & FD &
\textbf{0.212} & 0.645 & \textbf{0.0289} &
0.449 & 0.803 &
$1.324 \pm 0.933$ \\

Evolved on $F_{\sqrt{2}}$ & FD &
\underline{0.181} & \underline{0.834} & \underline{0.0250} &
0.426 & 0.770 &
$0.007 \pm 0.009$ \\

Evolved on w-Inter & GenS &
0.100 & 0.516 & 0.0110 &
\textbf{0.521} & 0.858 &
$0.012 \pm 0.005$ \\

PSFR: Evolved on Incl  & FD &
0.168 & \textbf{0.866} & 0.0217 &
\underline{0.489} & \underline{0.884} &
$0.050 \pm 0.022$ \\

\bottomrule
\end{tabular}
\end{table}

\textbf{PSFR objectives.} We evaluate the query-agnostic keyframe selector
independently of downstream MLLM reasoning using frame-level oracle annotations
(Table~\ref{tab:evolved_kfs_objectives}) under a $K=16$ frame budget. For the FindingDory dataset, we use its training split for keyframe-selector program evolution.
Intersection measures precision with worst-case aggregation over valid
groundings, Inclusion measures evidence coverage, and $F_{\sqrt2}$ balances
them while favoring coverage. We also report GenS weighted intersection and
CPU runtime over 100 validation FindingDory videos.

Compared with uniform, hand-crafted, and grid-searched selectors, program
evolution exposes a clear objective-dependent trade-off. Intersection gives
the best FindingDory precision but is slow and loses coverage;
$F_{\sqrt2}$ is more balanced; and the GenS-optimized program performs best on
its source protocol but transfers poorly. Inclusion yields the highest
FindingDory coverage, remains competitive on GenS, and has low CPU cost. We
therefore use it as PSFR, prioritizing retrieval of at least one valid
observation for downstream QA.

\textbf{Real-robot transfer.}
We test transfer to 32 minutes of real robot video: three traversals of the same street, recorded with an Intel RealSense D455 camera installed on an AgileX Scout 2.0 mobile robot. 
The 163 manually verified textual and binary questions in the Demo Outdoor Video QA Benchmark cover attributes, relative positions, actions, temporal order, and counting. Short-range questions describe a single observation supported by one evidence interval; medium-range questions compare encounters using $2.06\pm0.24$ intervals separated by mean gaps of $7.01\pm6.43$ minutes; and long-range questions combine distant events using $3.37\pm0.67$ intervals with mean gaps of $9.73\pm4.63$ minutes. Landmark-referenced annotations avoid explicit timestamps.
Annotators first selected and described frames or regions, grouped observations by subject or event, and marked distinctive landmarks. Candidate short-range questions were generated from these local facts, while medium- and long-range questions were constructed by combining facts across groups. Every answer was linked to supporting intervals, and all questions were checked for duplicate wording, valid timestamps, and consistency with the saved frames. The repeated street structure tests whether graph relations can disambiguate similar objects encountered at different landmarks and times.

With GPT-5-Mini and the same 32-frame budget, FocusGraph improves judge scores
at every temporal range and exact match on short and medium questions
(Table~\ref{tab:outdoor-diagnostic-qa}); long-range exact match remains the main
limitation.

\begin{table}[t]
\centering
\tiny
\caption{Demo Outdoor Video QA Benchmark results. Score is
the mean judge score (1--5), and EM is exact match (\%).}
\label{tab:outdoor-diagnostic-qa}
\setlength{\tabcolsep}{2pt}
\renewcommand{\arraystretch}{1.05}
\begin{tabular}{lcccccc}
\toprule
& \multicolumn{2}{c}{Short} & \multicolumn{2}{c}{Medium} & \multicolumn{2}{c}{Long} \\
\cmidrule(lr){2-3}\cmidrule(lr){4-5}\cmidrule(lr){6-7}
Questions & \multicolumn{2}{c}{46} & \multicolumn{2}{c}{52} & \multicolumn{2}{c}{65} \\
\midrule
Method & Score & EM & Score & EM & Score & EM \\
\midrule
\shortstack{Uniform} & 2.000 & 10.9 & 1.865 & 15.4 & 2.308 & \textbf{12.3} \\
FocusGraph & \textbf{3.717} & \textbf{45.7} & \textbf{2.673} & \textbf{28.8} & \textbf{2.508} & 10.8 \\
\bottomrule
\end{tabular}
\end{table}

\section{Conclusion}

In conclusion, we have presented FocusGraph, a novel framework that addresses the challenges of long-video question answering in embodied environments. By combining query-conditioned frame selection with a textual graph-based scene representation, our approach efficiently captures object interactions and temporal dependencies over extended horizons. This allows for accurate reasoning without the computational burden of processing every frame. Extensive evaluations on egocentric LVQA benchmarks demonstrate that FocusGraph achieves state-of-the-art accuracy while significantly improving inference efficiency. Our work underscores the value of structured, selective temporal abstraction in embodied video understanding and opens new avenues for scalable, intelligent agents capable of reasoning over complex, long-horizon visual experiences. Additionally, we demonstrate that the selection of semantically relevant video episodes can be effectively decoupled from keyframe extraction, which is driven by low-level visual properties, enabling a principled separation between semantic reasoning and visual redundancy reduction.

\section*{Acknowledgment}

GPT-6~\cite{gpt6} assisted with generating candidate benchmark questions from manual video annotations for Demo Outdoor Video QA Benchmark and polishing the paper text. The authors reviewed and revised the content and take responsibility for the final version.

{
    \small
    \bibliographystyle{IEEEtran}
    \bibliography{main}
}

\end{document}